\documentclass[accepted]{uai2026} %
\usepackage[american]{babel}
\usepackage{amsmath,amsfonts,bm,mathtools, amsthm}

\theoremstyle{definition}

\newtheorem*{example*}{Example}

\def\eqref#1{equation~\ref{#1}}

\def\1{\bm{1}}

\def\rx{{\textnormal{x}}}
\def\ry{{\textnormal{y}}}

\def\rvtheta{{\mathbf{\theta}}}

\def\rvx{{\mathbf{x}}}

\def\vtheta{{\bm{\theta}}}

\def\vu{{\bm{u}}}

\def\vx{{\bm{x}}}

\DeclareMathAlphabet{\mathsfit}{\encodingdefault}{\sfdefault}{m}{sl}
\SetMathAlphabet{\mathsfit}{bold}{\encodingdefault}{\sfdefault}{bx}{n}

\usepackage{natbib} %
\usepackage{mathtools}
\usepackage{minitoc}
\usepackage{hyperref}
\usepackage{url}
\usepackage{wrapfig}
\usepackage{subcaption}
\usepackage{algorithm}
\usepackage{booktabs}
\usepackage{algpseudocode}
\usepackage{siunitx}
\usepackage{bbm}
\usepackage[dvipsnames]{xcolor}

\usepackage{amssymb,amsmath,amsfonts}
\usepackage{xfrac}
\usepackage{tikz} %
\usetikzlibrary{positioning, arrows, calc, fit, shapes.geometric, shapes.misc, shadows, decorations.pathmorphing, decorations.pathreplacing, snakes, backgrounds}
\pgfdeclarelayer{background}
\pgfdeclarelayer{foreground}
\pgfsetlayers{background,main,foreground}

\definecolor{SafeRed}{RGB}{255, 0, 0}

\hypersetup{
  colorlinks = true,
  linkcolor  = ForestGreen,
  citecolor  = MidnightBlue,
  urlcolor   = BrickRed
}
\usepackage{enumitem}
\usepackage{graphicx}
\graphicspath{{figures/}{uai2026-template/figures/}}
\usepackage{multirow}
\usepackage[capitalize,noabbrev]{cleveref}
\usepackage{newtxmath}

\usepackage[createShortEnv,conf={no link to proof, text link},commandRef=Cref]{proof-at-the-end}

\makeatletter
\crefname{section}{\S\@gobble}{\S\@gobble}
\crefname{subsection}{\S\@gobble}{\S\@gobble}
\crefname{proposition}{Prop.}{Props.}
\crefname{figure}{Fig.}{Figs.}
\renewcommand{\eqref}[1]{(\ref{#1})}

\makeatother
\usepackage{xspace}
\newcommand{\ourMethod}{\textsc{ERRLESS}\xspace}
\newcommand{\ourMethodExtended}{\textsc{ERRLESS} (\underline{E}ntropy-\underline{R}egularized \underline{R}einforcement \underline{L}earning for \underline{E}xpression \underline{S}tructure \underline{S}ampling)}
\newcommand{\eg}{\textit{e.g.}}
\newcommand{\ie}{\textit{i.e.}}

\newcommand{\T}{\textnormal{T}} %
\newcommand{\D}{\mathcal{D}} %
\newcommand{\dd}{\mathrm{d}}

\newcommand{\TSpace}{\mathcal{T}_{\Sigma}}
\newcommand{\TSpaceValid}{\mathcal{T}^\circ}

\newcommand{\IC}{I_{\mathcal{C}}(\T)}
\newcommand{\IV}{I_{\mathcal{V}}(\T)}

\usepackage[most]{tcolorbox}

\makeatletter
    \titleformat*{\section}{\raggedright\large\bfseries\MakeUppercase}
    \titleformat*{\subsection}{\raggedright\bfseries\MakeUppercase}
    \titlespacing{\section}{\z@}{*2}{*1}
    \titlespacing{\subsection}{\z@}{*2}{*1}
    \titlespacing{\subsubsection}{\z@}{*2}{*1}
    \titlespacing*{\paragraph}{0pt}{0pt}{0.6em}
\makeatother

\tcbset{
  myexamplebox/.style={
    enhanced,
    colback=gray!10!black!5,   %
    colframe=black!70,         %
    boxrule=1pt,               %
    arc=8pt,                   %
    left=6pt, right=6pt, top=6pt, bottom=6pt,
    drop shadow={shadow xshift=2pt, shadow yshift=-2pt, opacity=0.15}
  }
}

\title{Bayesian Symbolic Regression with Entropic Reinforcement Learning}

\author[1,2]{\href{mailto:<oussama.boussif@mila.quebec>?Subject=About your UAI 2026 paper, ERRLESS}{Oussama~Boussif}{}}
\author[3]{Mohammed~Mahfoud}
\author[4]{Younesse~Kaddar}
\author[1,2]{Moksh~Jain}
\author[5]{Sida~Li}
\author[6]{Damiano~Fornasiere}
\author[1,2]{Xiaoyin~Chen}
\author[1,2,6,7]{Yoshua~Bengio}
\author[7,8]{Esmeralda~S.~Whitammer}

\affil[1]{Mila – Québec AI Institute}
\affil[2]{%
    Université de Montréal
}
\affil[3]{%
    Independent
}
\affil[4]{%
    University of Oxford%
}
\affil[5]{%
    University of Chicago%
}
\affil[6]{%
    LawZero
}
\affil[7]{%
    CIFAR Fellow
  }
\affil[8]{%
    University of Edinburgh%
}

\begin{document}
\maketitle

\begin{abstract}
  Symbolic regression is the problem of finding an algebraic expression describing a stochastic dependence of a target variable on a set of inputs. Unlike forms of regression that fit parameters assuming a fixed model structure, symbolic regression is a search problem over the space of expressions, represented, for example, as abstract syntax trees using a library of operators. Symbolic regression is typically used in settings with limited, noisy data in the natural sciences. However, searching for a single best-fitting expression fails to capture the epistemic uncertainty about the expression, which motivates a Bayesian perspective that enables uncertainty quantification and specification of natural priors to constrain the search space. In this work, we propose \ourMethodExtended \footnote{The code is available at \url{https://github.com/jaggbow/ERRLESS}}, a scalable approach for sampling from the posterior distribution over expressions given data using maximum-entropy reinforcement learning. \ourMethod~learns a neural policy that constructs expressions sequentially by building up their abstract syntax trees. At convergence, the policy samples expressions from the posterior. At test time, expressions can be sampled by rollouts of this policy. We demonstrate that \ourMethod~achieves competitive results on the Feynman benchmark while producing short and interpretable expressions. Additionally, we demonstrate that the mean of the posterior predictive approximated by \ourMethod~achieves a higher coefficient of determination ($R^2$) compared to an SMC baseline, which shows the value of the Bayesian perspective in symbolic regression.
\end{abstract}

\section{Introduction}

\label{intro}

Symbolic regression (SR) is the problem of searching over a space of compositional algebraic expressions, using a certain library of primitive operators, to find a function that most closely maps the inputs observed in a dataset to their corresponding outputs. SR is a common problem in the natural sciences, where datasets are small and noisy, domain priors constrain plausible formulas, and interpretability is important \citep{bongard2007automated, schmidt2009distilling, Udrescu2020}. Most existing algorithms for SR \citep{petersen2019deep, biggio2021neural, mundhenk2021symbolic, PhySO_RL_DA, kamienny2023deep} have the goal of finding a Pareto front of expressions that trade off complexity with fit. With limited data, such a point estimate can be unreliable and hides the uncertainty about the expression arising from the noise and scarcity of data. 

\looseness=-1
The need to model uncertainty in SR has been recognized in the literature and addressed by a Bayesian perspective on the problem \citep{jin2019bayesian, guimera2025bayesian}. In this view, one posits a (structured) prior over expression structures and parameters and a model of observation noise. A dataset of (input, output) pairs then induces a posterior distribution over expressions, and the aim of Bayesian SR is to sample from this posterior. Previous Bayesian SR methods use reversible-jump Markov chain Monte Carlo (MCMC)~\citep{green1995reversible} or sequential Monte Carlo (SMC)~\citep{bomarito2025bayesian,guimera2025bayesian}, but these methods rely on handcrafted proposal distributions and can be costly to scale. In this paper, we instead seek an approach that amortizes the sampling process using a neural network.

\looseness=-1
We formulate the construction of an expression as a sequential decision-making problem, turning the Bayesian SR task into the reinforcement learning (RL) problem of training a policy to sample expressions from the posterior. To achieve unbiased sampling at convergence, we train this policy using a maximum-entropy RL objective with the unnormalized posterior log-density as the reward (\Cref{rl}). The resulting system, which we call \ourMethodExtended, is capable of learning a policy that samples expression structures and their real-valued parameters; once trained, the policy can be sampled to produce approximate posterior samples efficiently.

Successfully applying entropy-regularized RL to Bayesian SR requires careful design choices. \ourMethod uses a generation process that constructs expression syntax trees in a bottom-up (postorder) manner to allow effective imposition of constraints, imposes structural priors to avoid ill-formed, redundant, or unlikely subexpressions, and can incorporate additional constraints to restrict the search space to forbid dimensionally incompatible compositions of physical units (\Cref{space_of_expressions}). The parametrization and training of the policy similarly require appropriate design of neural architectures and off-policy training schemes (\Cref{design}). Finally, unlike previous Monte Carlo-based methods, \ourMethod amortizes sampling of both expression structures and parameter values into a single neural policy and avoids explicit per-candidate constant fitting. 

On the Feynman Symbolic Regression Database~\citep{Udrescu2020}, \ourMethod achieves a competitive prediction accuracy and produces posterior samples that improve predictions when data are few and noisy. On synthetic data, the posterior predictive mean of \ourMethod is better than PySIPS \citep{bomarito2025bayesian}, an SMC baseline on the noisiest setting, showing the value of modeling uncertainty over expressions with a Bayesian view.

We summarize our contributions as follows:
\begin{itemize}[left=0pt,nosep]
\item We design a novel generation process (environment) for symbolic regression that enables bottom-up expression construction, incorporates dimensional analysis, and imposes structural constraints.
\item We formulate Bayesian symbolic regression, including inference of scalar parameters in expressions, as an end-to-end policy-learning problem within this environment.
\item We demonstrate that \ourMethod achieves competitive prediction accuracy compared to state-of-the-art approaches on the Feynman Symbolic Regression Database while producing short expressions.
\item We show that our approach captures the posterior distribution effectively, facilitating downstream applications in settings with scarce and noisy data.
\end{itemize}
    
\begin{figure*}[t]
\vspace*{-1em}
    \centering
    \resizebox{\linewidth}{!}{\input{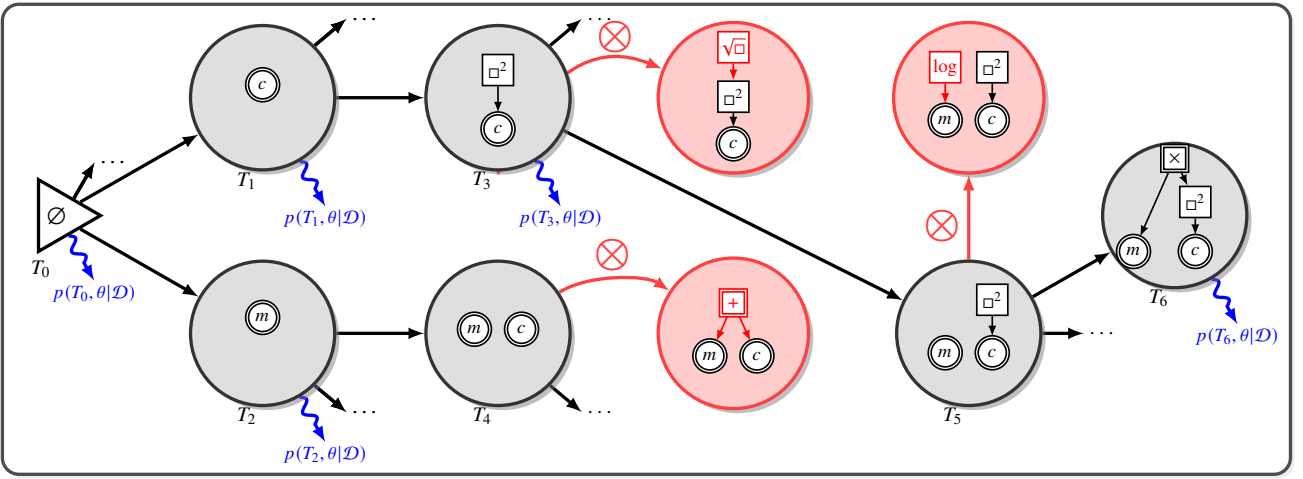}}
    \caption{\textbf{Generation process for the expression $E = mc^2$}. Valid states are shown in gray, while invalid states have pink fill and red outlines. Black arrows indicate valid transitions, and red arrows indicate invalid transitions. Leaf nodes are represented by double-outlined circles, unary operators by single-outlined squares, and binary operators by double-outlined squares. Wiggly arrows denote states that can terminate and transition to the terminal state.}
    \label{fig:method}
\end{figure*}

\section{Problem setting}
\label{prelim}

In this section, we describe the formal setting for Bayesian SR. \Cref{space_of_expressions} describes the search space and \Cref{bsr} formulates the sampling problem.

\paragraph{Notation.} Scalar (resp. vector) random variables, or variables whose type has not been specified, are denoted $\rx$ (resp. $\rvx$), and their realization is denoted $x$ (resp. $\vx$). The probability distribution of a discrete random variable $\rx$ is denoted $P_\rx$ (in uppercase) and that of a continuous random variable, or a variable whose type has not been specified, is denoted $p_\rx$ (in lowercase). Let $\mathbb{N}$ (resp. $\mathbb{R}$) be the set of natural (resp. real) numbers, and $\mathbb{N}_{>0} \coloneq \mathbb{N} \setminus \{0\}$ (resp. $\mathbb{R}_{>0} \coloneq \{x \in \mathbb{R} \mid x > 0\}$).
For $n\in\mathbb{N}$, we denote the set of integers $\{1,\dots,n\}$ (or the empty set when $n = 0$) as $[n]$.

For a set $S$, we denote by $|S|$ its cardinality and by $S^\ast \coloneq \bigcup_{n \in \mathbb{N}} S^n$ the set of all finite sequences (strings) over $S$.

\subsection{The space of expressions}
\label{space_of_expressions}

\paragraph{Expression trees.} We fix a vocabulary $\Sigma \coloneq \mathcal{V} \sqcup \mathcal{C} \sqcup \mathcal{O}$, where $\mathcal{V} \coloneq \{v_i\}_{i \in [D]}$ is a set of \emph{variable symbols}, $\mathcal{C} \coloneq \{c_k\}_{k \in [K]}$ is a set of \emph{constant symbols}, and $\mathcal{O} \coloneq \{g_j\}_{j \in [M]}$ is a set of \emph{operator symbols} ($\mathcal{O}$ is referred to as the \emph{operator library}).
Each operator $g\in \mathcal{O}$ is equipped with an \emph{arity} $r_g\in\mathbb{N}_{>0}$ and a \emph{semantic function} $\phi_g\colon \mathcal{X}_g \to \mathbb{R}$, where $\mathcal{X}_g\subseteq\mathbb{R}^{r_g}$. For example, the operator `$+$' is binary and $\phi_+\colon \mathbb{R}^2 \to \mathbb{R}, (x,y)\mapsto x+y$; the operator `$\log$' is unary and $\phi_{\log}\colon\mathbb{R}_{>0}\to\mathbb{R}, x\mapsto\log x$.

An \emph{expression tree} $\T$ is a finite, rooted, ordered tree where each leaf is labeled with a symbol from $\mathcal{V} \sqcup \mathcal{C}$, and each internal node with $r$ children is labeled with a $g \in \mathcal{O}$ such that $r_g = r$. Let $\IC\subseteq [K]$ (resp. $\IV \subseteq [D]$) be the set of indices of constants (resp. variables) appearing in $\T$. The constants $\{c_k\}_{k \in \IC}$ are thought of as symbolic placeholders: a \emph{constant assignment} is a vector $\vtheta \in \mathbb{R}^K$ specifying numerical values for the symbols at indices $\IC$.

Let $\TSpace$ be the space of expression trees over $\Sigma$.

\paragraph{Evaluation.} Let $\T\in\TSpace$. Given a constant assignment $\vtheta\coloneq(\theta_k)_{k \in [K]}\in\mathbb{R}^{K}$, one can define a partial \emph{evaluation function} $f_{\T, \vtheta}\colon \mathbb{R}^{D} \to \mathbb{R}$ for $\T$ on every input $\vx\coloneq(x_i)_{i \in [D]}\in\mathbb{R}^{D}$ recursively as follows:
\begin{itemize}[left=0pt,nosep]
    \item each leaf of $\T$ labeled $v_i \in \mathcal{V}$ (resp. $c_k \in \mathcal{C}$), where $i \in \IV$ (resp. $k \in \IC$), evaluates to $x_i$ (resp. $\theta_k$);
    \item each internal node of $\T$ labeled with operator $g$ having children that evaluate to $a_1, \ldots, a_{r_g}$ evaluates to $\phi_g(a_1, \ldots, a_{r_g})$ if $(a_1, \ldots, a_{r_g}) \in \mathcal{X}_g$, otherwise the evaluation is undefined.
\end{itemize}

\paragraph{Sequence representation.} An expression tree $\T$ can be canonically represented as a sequence $\mathcal{W}(\T) \in \Sigma^\ast$ by listing the labels of its nodes in postorder traversal, a representation known as reverse Polish notation \citep{Lukasiewicz1929}. For example, the expression $\sin(v_1+c_1\times v_2)$ is represented by the sequence $(v_1,c_1,v_2,\times,+,\sin)$. (It is well-known that such an expression can be evaluated, when real numbers are substituted for the variables and constants, by traversing the sequence from left to right and maintaining a stack.) 

\paragraph{Unit constraints.} One can additionally constrain the space of trees by modifying the above definitions to include dimension information. We assume that each variable $v_i$ has an associated physical unit (\eg, meters, kilograms), represented as a vector $\vu_i \in \mathbb{R}^\ell$ (called \emph{unit vector}), where $\ell \coloneq 7$ is the number of base units. Each coordinate of $\vu_i$ represents the exponent of its corresponding unit, \eg, velocity has units \si{m.s^{-1}}, represented as $(1, 0, -1, 0, 0, 0, 0)$. Dimensionless variables are represented by the zero vector $(0, \ldots, 0) \in \mathbb{R}^\ell$. All constants $c_k$ are dimensionless. 

Every operator $g$ is assumed to have constraints on the unit vectors of its arguments, as well as a mapping from the unit vectors of the arguments to the unit vector of its output. Formally, every $g$ has a \emph{unit assignment function}, a partial function $\mathcal{U}_g:(\mathbb{R}^\ell)^{r_g} \rightarrow \mathbb{R}^\ell$ that returns the unit vector of the operator's output if the operator can be applied to inputs with the given units (and is undefined otherwise). 
Let $\TSpaceValid\subseteq\TSpace$ be the set of those expression trees which are dimensionally valid: when evaluating bottom-up, $\mathcal{U}_g$ is never undefined when evaluated on the unit vectors of the children of any internal node. See \cref{unit_constraints} for the assignment functions of operators in the library we use.

\begin{tcolorbox}[myexamplebox]
\begin{example*}
Consider the following operators applied to $\vu^{\text{vel}} \coloneq (1, 0, -1, 0, 0, 0, 0)$, $\vu^{\text{time}} \coloneq (0, 0, 1, 0, 0, 0, 0)$, and $\vu^{\text{angle}} \coloneq (0, 0, 0, 0, 0, 0, 0)$:
\begin{itemize}[left=0pt,nosep]
    \item $+$: $\mathcal{U}_+(\vu^{\text{vel}}, \vu^{\text{time}})$ is undefined (incompatible units: time and velocity cannot be added), but $\mathcal{U}_+(\vu^{\text{vel}}, \vu^{\text{vel}}) = \vu^{\text{vel}}$ (adding two velocities gives a velocity);
    \item $\times$: $\mathcal{U}_\times(\vu^{\text{vel}}, \vu^{\text{time}}) = (1, 0, 0, 0, 0, 0, 0)$ (multiplying velocity and time gives a distance);
    \item $\sin$: $\mathcal{U}_{\sin}(\vu^{\text{angle}})= \vu^{\text{angle}}$ (sine of a scalar is a scalar), but $\mathcal{U}_{\sin}(\vu^{\text{vel}})$ is undefined.
\end{itemize}
\end{example*}
\end{tcolorbox}

\subsection{Bayesian symbolic regression} 
\label{bsr}

We fix a prior probability distribution $P(\T)$ over the set of valid expression trees $\TSpaceValid$ (\Cref{design} describes the choice of prior) and a conditional prior distribution $p(\vtheta,\sigma \mid\T)$ over the space $\mathbb{R}^{K}$ of constant assignments and the noise variance.

Let $\{(\rvx_i,\ry_i)\}_{i=1}^N$ be a collection of random variables such that, for every $i \in [N]$, $(\rvx_i, \ry_i)$ takes values in $\mathbb{R}^D \times \mathbb{R}$, and each realization $\vx_i \in \mathbb{R}^D$ of $\rvx_i$ defines the following conditional distribution:
\begin{equation}\label{conditional-rv}
    \ry_i \mid \left(\rvx_i=\vx_i, \T, \rvtheta, \sigma \right) \sim \mathscr{N}\big(f_{\T,\vtheta}(\vx_i), \sigma^2\big),
\end{equation}
where $\sigma^2 \in\mathbb{R}_{> 0}$ is a noise variance and the $\ry_i$'s are conditionally independent given the $\rvx_i$'s. That is, $\ry_i$ is the evaluation of the expression given by $\T,\vtheta$ with inputs $\vx_i$, with added Gaussian noise.

We observe a dataset $\D\coloneq\{(\vx_i, y_i)\}_{i=1}^N$. Assuming $\vx_i$ are independent of $\T$ and $\vtheta$ and $\sigma$, the posterior of $(\T, \vtheta,\sigma)$ given $\D$ satisfies the following relation:
\begin{align}
    &p(\T,\vtheta,\sigma \mid \D) \nonumber
    \\&\propto p(\T,\vtheta,\sigma)  \cdot p(\D \mid \T,\vtheta,\sigma) \nonumber \\
    &\propto p(\T,\vtheta,\sigma) \prod_{i=1}^N p(y_i \mid \vx_i, \T, \vtheta,\sigma) \underbrace{p(\vx_i \mid \T, \vtheta, \sigma)}_{= \, p(\vx_i)} \nonumber \\
    &\propto P(\T)p(\vtheta,\sigma\mid \T) \prod_{i=1}^N 
    \mathcal{N}(y_i; f_{\T,\vtheta}(\vx_i),\sigma^2)
    \label{posterior}
\end{align}
The objective of Bayesian symbolic regression is to sample from this posterior distribution. The next sections introduce the framework we use to approximate it in practice.

\section{Methodology}
\label{methodology}

We introduce \ourMethodExtended, a scalable approach for Bayesian symbolic regression, illustrated in \cref{fig:method}. \cref{bottomup} introduces a sequential decision-making process for sampling expression trees that enforces constraints during generation, \cref{rl} describes learning objectives for decision-making policies, and \cref{design} describes the policy architecture used to sample expressions.

For concreteness, we specify the set of operators we will use in our experiments: the unary operators 
$\big\{\sin, \cos, \log, \exp, \square^2, \sqrt{\square}, -\square\big\}$ and the binary operators
$\big\{+, -, /, \times \big\}$ (see \Cref{unit_constraints} for details). However, the algorithm described below is not restricted to this particular choice.

\subsection{Bottom-up generation}
\label{bottomup}

Most approaches in the literature \citep{petersen2019deep,li2023gfn} employ \emph{top-down generation}, where internal nodes (\ie, operators) are sampled before leaf nodes (input variables and constants). With top-down generation, the intermediate expression at each step contains `holes' to be filled. As a result, the expression can only be evaluated, and the posterior density computed, at the end of the generation. Moreover, with top-down generation, enforcing physical unit constraints is inefficient: since the operands may not yet be specified at intermediate steps, assigning units requires traversing the full depth of the tree once the units have been determined.

In contrast, \emph{bottom-up generation} offers two main advantages. First, some intermediate states are valid and complete expressions. Second, because leaf nodes (and therefore operands) are specified from the outset, newly added operators can be constrained to be compatible with the known physical units of the operands.

\paragraph{Sequential generation of expression trees.}

Recall that an expression tree $\T$ is uniquely represented by its postorder representation $\mathcal{W}(\T) \in \Sigma^\ast$. Therefore, generating an expression tree is equivalent to generating its sequence representation by appending one symbol at a time from left to right. Because all operator nodes appear after their arguments in postorder traversal, the dimensional validity and arity constraints correspond to restrictions on continuations of partial sequences. The problem of modeling a probability distribution over expression trees is thus reduced to one of autoregressive sequence modeling.

We define the token alphabet $\mathcal{A} \coloneq \Sigma \sqcup \{\top\}$, where $\top$ is a special symbol marking sequence termination. A distribution $\pi$ over the set of trees $\TSpaceValid$ is equivalent to a distribution over $\mathcal{A}^\ast$:
\begin{equation}
    \pi(w_1 \dots w_n \top) = \left( \prod_{i=1}^{n} \pi(w_i \mid w_{<i}) \right) \pi(\top \mid w_{1 \dots n}),
    \label{eq:autoreg}
\end{equation}
where $w_i\in\mathcal{A}$ and the support of each next-symbol distribution respects the unit and arity constraints, as well as the restriction that generation of $\top$ is permitted only after a sequence that represents a complete expression tree. (We must also make the assumption that every partial sequence can be continued to at least one sequence ending in $\top$, which holds with the library of operators we consider.)

\begin{figure}
    \centering
  \vspace{-.5cm}
  \includegraphics[width=.99\linewidth]{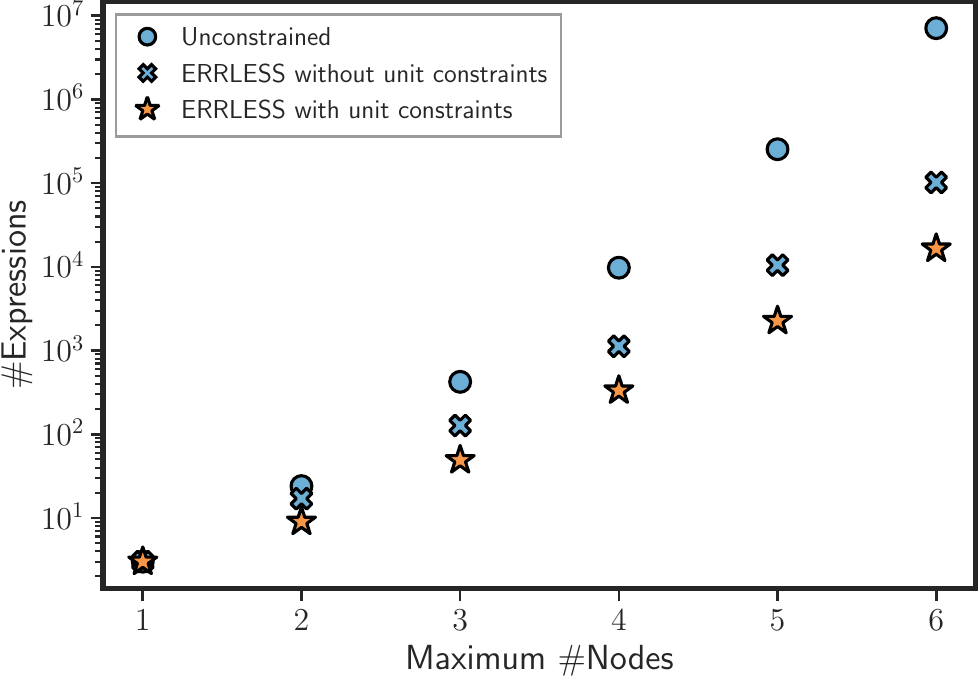}
  \caption{Growth of search space size in maximum number of nodes (in $\log$-scale).}
  \vspace*{-.4cm}
  \label{search_space}
\end{figure}

\paragraph{Length and redundancy constraints.} We remark that constraints on the number of nodes in the expression tree $\T$ can also be expressed as restrictions on the support of the next-token distributions of $\mathcal{W}(\T)$ in \eqref{eq:autoreg}. In addition, structural constraints in the target distribution that disallow ill-formed, redundant, or unlikely compositions can also be expressed as restrictions on next-token distributions. The constraints we impose prohibit: (i) composing functions with their inverses (\eg, $\sqrt{\square}^2$), (ii) nesting trigonometric functions (\eg, $\sin(\cos(\square))$), (iii) nesting exponentials (\eg, $e^{e^{\square}}$), and (iv) applying unary operators directly to constants. \cref{search_space} shows that our construction rules and constraints make the search space much smaller and thus more efficient to explore.

\paragraph{Samplers of expressions and parameters as policies.}
A distribution $\pi$ over sequences of the form \eqref{eq:autoreg} respecting the imposed constraints, together with a conditional distribution $\pi(\vtheta,\sigma\mid\T)$ over constant assignments and noise given a tree, define a joint distribution over $(\T,\vtheta,\sigma)$. In the next section, we will describe how this autoregressively factorized distribution can be trained to sample the Bayesian posterior defined in \cref{bsr} using reinforcement learning methods.

\subsection{Maximum-entropy RL training of expression samplers}
\label{rl}

Above, we have identified a distribution $\pi$ over tuples $(\T,\vtheta,\sigma)$ with a distribution over sequence representations and a conditional distribution over constant assignments and noise:
\[
    \pi(\T,\vtheta,\sigma)
    =\pi(\T)\pi(\vtheta,\sigma\mid\T)
    =\pi(\mathcal{W}(\T)\top)\pi(\vtheta,\sigma\mid\T),
\]
where $\pi(\mathcal{W}(\T)\top)$ has an autoregressive factorization \eqref{eq:autoreg}. Suppose that $\pi$ is a parametric model $\pi_\varphi$, which can be evaluated to obtain next-token logits for sequential generation of $\mathcal{W}(\T)\top$ and the parameters of the distribution over $\vtheta$ given $\T$ (for example, the mean and covariance of a Gaussian from which $\vtheta$ is sampled). Our goal is to fit $\varphi$ so that $\pi_\varphi(\T,\vtheta,\sigma)$ equals the posterior $p(\T,\vtheta,\sigma\mid \D)$ defined in \eqref{posterior}, which is given as an unnormalized probability density function.

Define
\[
    R(\T,\vtheta,\sigma)
    =
    \log p(\T,\vtheta,\sigma) + \sum_{i=1}^N \log\mathcal{N}(y_i; f_{\T,\vtheta}(\vx_i),\sigma^2),
\]
so that $p(\T,\vtheta,\sigma\mid \D)\propto\exp(R(\T,\vtheta,\sigma))$ (cf.~\eqref{posterior}). When $R$ is thought of as a \emph{reward} provided to a sampler that generates $\T$, $\vtheta$ and $\sigma$ following the conditional distributions of $\pi_\varphi$, training $\pi_\varphi$ to maximize its expected reward $\mathbb{E}_{(\T,\vtheta,\sigma)\sim\pi_\varphi} [R(\T,\vtheta,\sigma)]$ is equivalent to minimizing cross-entropy between $\pi_\varphi$ and the posterior, which is achieved by sampling the posterior mode. However, one can instead consider the \emph{entropy-regularized} problem
\begin{equation}
    \max_\varphi\left[ \mathbb{E}_{(\T,\vtheta,\sigma)\sim\pi_\varphi} [R(\T,\vtheta,\sigma)] + \mathcal{H}[\pi_\varphi]\right],
    \label{eq:maxent}
\end{equation}
where $\mathcal{H}[\pi_\varphi]$ is the entropy of the modeled distribution. A key property of entropy-regularized, or maximum-entropy RL \citep{haarnoja2018soft,eysenbach2022maximumentropyrlprovably} is that the solution to \eqref{eq:maxent} minimizes KL divergence between $\pi_\varphi$ and the distribution with density proportional to $\exp(R(\T,\vtheta,\sigma))$, \ie, the posterior. This property has been exploited in various instances of learning to sample by sequential decision-making \citep{deleu2024discrete}, and efficient off-policy training algorithms for solving \eqref{eq:maxent} have been proposed.

\paragraph{Training objective.} One such off-policy objective is the \emph{trajectory balance} (TB) objective \citep{malkin2022trajectory}, a core GFlowNet \citep{bengio2021flow} loss denoted by $\mathcal{L}_{\text{TB}}$, which is a special case of a path consistency learning objective \citep{nachum2017bridging} in deterministic environments with sparse terminal rewards.
TB requires additionally learning a scalar parameter $\log Z_\varphi$ (corresponding to the initial state's value function in reinforcement learning), which, at convergence, gives the log normalizing constant (the likelihood of $\D$). The objective associated with $(\T,\vtheta,\sigma)$ is:
\begin{equation}
    \mathcal{L}_{\text{TB}}(\T, \vtheta, \sigma;\varphi) \coloneq \big[\log Z_\varphi + \log \pi_\varphi(\T,\vtheta,\sigma) - R(\T, \vtheta,\sigma)\big]^2.
    \label{eq:tb}
\end{equation}
Because \eqref{eq:tb} can be minimized to $0$ for \emph{all} samples simultaneously, training algorithms can optimize this objective w.r.t. $\varphi$ over $(\T,\vtheta,\sigma)$ sampled from some behavior policy that does not necessarily coincide with the current state of $\pi_\varphi$ itself. We describe our off-policy training choices in \Cref{design} and ablate the training objective in \cref{sec:ablation_objective}. (Because the reward can equal $0$, we apply smoothing to prevent $\log 0$ in the loss; see \Cref{unigram}.) Assuming convergence of the on-policy estimate of the objective in \eqref{eq:maxent} to its optimal value (the normalizing constant), convergence in policy space is guaranteed by Theorem 3.1 of \cite{ICLR2025_f3efbcfe}.

\subsection{\ourMethod design choices}
\label{design}

\paragraph{Parametrization of the policy.} 

We represent each partial expression as a sequence of tokens and parameterize the policy with a transformer-like architecture~\citep{attention}. The transformer encodes the sequence into an embedding that is supplied to three prediction heads: (i) a head that produces the logits over the next action at each step of the construction process, for a sequence that has not terminated in $\top$, (ii) a head that outputs the means, covariances, and mixture weights of a Gaussian mixture model from which the values of the constants are sampled for a sequence that has terminated and (iii) a head that outputs the means, variances, and mixture weights of a mixture of log-normals from which the noise standard deviation $\sigma$ is sampled (see \Cref{exp_details} for more details). 

\paragraph{Prior and tempering.} 
We choose a unigram prior $P(\T)$ over expression trees (see \cref{unigram} for more details), a choice that we ablate in \cref{sec:ablation_prior}. We also impose length constraints and the redundancy constraints described in \cref{bottomup}. For the prior $p(\vtheta\mid \T)$ over constant assignments, we chose a Gaussian prior with mean $0$ and standard deviation of $10$ unless specified otherwise with a penalty for the number of constants used. The prior $p(\sigma)$ over $\sigma$ was chosen to be a wide half-Normal distribution with a scale parameter of $2000$ for the Feynman and Blackbox datasets, and a LogNormal distribution for the synthetic dataset.

\paragraph{Training policy.} To encourage exploration, we use off-policy training and use $\epsilon$-greedy exploration with annealed $\epsilon$ along with a prioritized replay buffer. See \cref{exp_details} for all relevant hyperparameters and training details. To stabilize the training objective, inspired by \citet{deleu2022bayesian}, we use a modified version of TB that clips the gradients of the squared loss, analogous to the Huber loss.

\section{Related works}
\label{related_works}

\textbf{Learning and Search-Based Symbolic Regression.} Deep symbolic regression~\citep[DSR;][]{petersen2019deep} trains an autoregressive recurrent neural network (RNN) policy via risk-seeking policy gradient algorithms. \citet{PhySO_RL_DA} uses the same learning algorithm but enforces constraints on physical units at each step of the generation. \ourMethod differs from both approaches by taking a Bayesian perspective using maximum entropy RL to train the policy and using a bottom-up generative process. ~\citet[CADSR;][]{bastiani2024complexity0aware} use the Bayesian information criterion (BIC) as a reward for their risk-seeking policy gradient algorithm which approximates the posterior over expressions with Laplace approximation, but unlike \ourMethod, their learned policy does not sample proportionally to the posterior.  ~\citet{mundhenk2021symbolic} combine DSR with genetic programming, resulting in improved exploration and recovery of benchmark formulas. ~\citet[PySR;][]{cranmer2023interpretable} takes a purely evolutionary approach, combining population-based search with gradient-based parameter optimization to establish a widely-adopted, practical baseline that avoids neural policies entirely. Beyond RL and evolutionary approaches, NeSymReS~\citep{biggio2021neural} is a transformer pre-trained on an equation corpus, yielding zero-shot generalization across diverse symbolic regression tasks. \citet{kamienny2023deep} integrates a pre-trained model within Monte Carlo Tree Search (MCTS) whereas ~\citet[SPL;][]{sun2022symbolic} use an MCTS-only approach with no neural policy. Unlike these approaches, \ourMethod does not rely on existing datasets and adopts a Bayesian perspective on symbolic regression. 

\textbf{Bayesian symbolic regression and probabilistic modeling.}
The method Bayesian Symbolic Regression \citep[BSR;][]{jin2019bayesian} uses reversible-jump MCMC~\citep{green1995reversible} to sample expressions from the posterior distribution. \citet{bomarito2025bayesian} replaces MCMC with sequential Monte Carlo (SMC). Finally, \citet{guimera2025bayesian} provides a statistical physics perspective on Bayesian symbolic regression. Unlike existing BSR methods that rely on handcrafted proposal distributions and computationally intensive MCMC sampling, \ourMethod learns an amortized posterior sampler with maximum entropy reinforcement learning.

\paragraph{Sampling discrete structured posteriors with maximum-entropy reinforcement learning.} 
\ourMethod builds upon prior work on maximum entropy RL for sampling from discrete structured distributions~\citep{buesing2020approximate,bengio2021flow}. The trajectory balance objective~\citep{malkin2022trajectory}, equivalent to path consistency learning~\citep{nachum2017bridging}, has been used for posterior inference over decision trees~\citep{mahfoud2025learning}, causal models~\citep{deleu2022bayesian,deleu2023jointbayesianinferencegraphical}, phylogenetic trees~\citep{zhou2024phylogfnphylogeneticinferencegenerative}. GFN-SR~\citep{li2023gfn} uses trajectory balance to learn policies for sampling expressions on small synthetic benchmarks. \ourMethod deviates from GFN-SR by using a better generative process for constructing expressions, incorporating explicit priors, and evaluating on large-scale benchmarks complemented by improved training. 

\begin{table*}[t]
\vspace*{-1em}
\centering
  \caption{\textbf{Posterior predictive metrics on the synthetic dataset.} We report the accuracy of the mean of the posterior predictive ($R^2_{\rm PP}$) as well as the negative log-likelihood (NLL) on the test set, with both methods scored by the same estimator (\cref{app:nll-estimator}): a pointwise mixture NLL with uniform weights over up to $1000$ posterior draws, where $\sigma$ is each draw's sampled value for \ourMethod and each particle's residual error on the training split for PySIPS. Test $R^2$ is the test-set $R^2$ of the single best expression found by each method. Median of $10$ independent runs. $^\dagger$: only 2 of 10 PySIPS runs yield a finite posterior-mean prediction on this problem. \label{tab:synthetic_results}}
\begin{tabular}{@{}l cc cc cc }
\toprule
Metric $\rightarrow$ & \multicolumn{2}{c}{\textbf{NLL} ($\downarrow$)} & \multicolumn{2}{c}{\textbf{Test $R^2$} ($\uparrow$)} & \multicolumn{2}{c}{\textbf{$R^2_{\mathrm{PP}}$} ($\uparrow$)} \\
\cmidrule(lr){2-3} \cmidrule(lr){4-5} \cmidrule(lr){6-7} 
 Expression $\downarrow$ Algorithm $\rightarrow$ & ERRLESS & PySIPS & ERRLESS & PySIPS & ERRLESS & PySIPS \\
\midrule
{$1/\sqrt{1-x^2}$} & $60.832$ & {$\bf -57.645$} & $0.101$ & {$\bf 0.532$} & {$\bf -0.232$} & $-8.9{\times}10^{53}$ \\
{$2.37x+3.02$} & $114.973$ & {$\bf -3.255$} & $0.969$ & {$\bf 0.974$} & {$\bf 0.760$} & $0.349$ \\
{$4.567\exp{x}+y$} & $183.971$ & {$\bf 111.125$} & $0.976$ & {$\bf 1.000$} & {$\bf 0.978$} & $-1.5{\times}10^{91}$ \\
{$\sin{(y+2.5x)}$} & $182.595$ & {$\bf -80.669$} & $-0.607$ & {$\bf 0.999$} & {$\bf -5.708$} & $-70.162$ \\
{$\sin{x}\cos{y}$} & $343.812$ & {$\bf -20.211$} & $-2.275$ & {$\bf 0.401$} & {$\bf -11.055$} & $-1.5{\times}10^{272}$ \\
{$\sqrt{x^2+y^2}$} & $210.603$ & {$\bf -43.306^\dagger$} & {$\bf 0.993$} & $0.815^\dagger$ & {$\bf -9.387$} & $-\infty^\dagger$ \\
{$x+\sin{(5.5x)}$} & $377.029$ & {$\bf -13.233$} & $-0.799$ & {$\bf 0.323$} & {$\bf -2.230$} & $-2.8{\times}10^{38}$ \\
\bottomrule
\end{tabular}
\vspace*{-0.5em}
\end{table*}

\begin{figure*}[t]
    \centering
    \includegraphics[width=.99\linewidth]{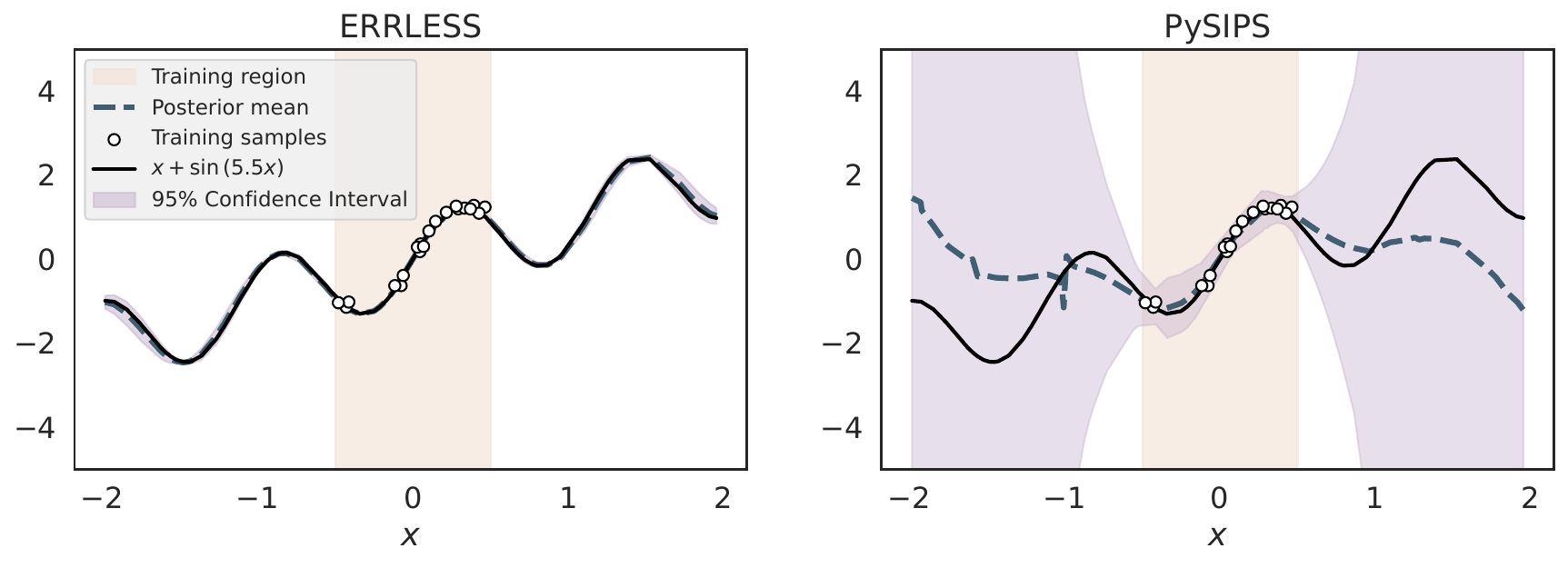}
    \vspace*{-1.5em}
    \caption{\textbf{Samples from the posterior}. \ourMethod and PySIPS are trained on noised values (white circles) of the ground-truth function $x+\sin(5.5x)$ (solid black line) at points in the training domain $[-0.5,0.5]$ (beige). The posterior mean (dashed blue line), that is, the mean of the individual posterior samples, fits the true function well on the beige interval in both cases, but \ourMethod extrapolates better outside the training region. We show in purple the 95\% credible intervals. For \ourMethod both the mean and credible intervals are computed using importance sampling. Results shown are from a single run. \Cref{tab:synthetic_results} reports medians over ten runs.}
    \label{fig:posterior_plot}
    \vspace*{-1em}
\end{figure*}
\begin{figure*}[t]
\vspace*{-0.5em}
    \centering
    \begin{subfigure}[b]{0.35\linewidth}
        \centering
        \includegraphics[width=\linewidth]{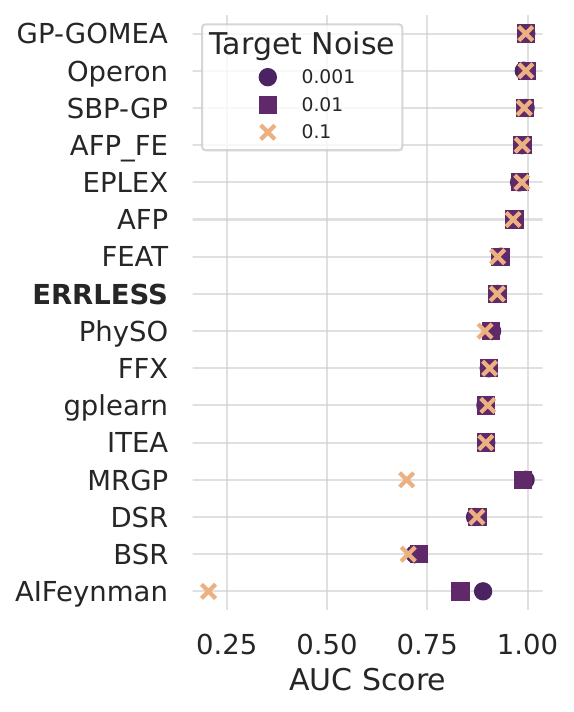}
        \caption{Robustness to target noise}
        \label{fig:feynman}
    \end{subfigure}
    \hfill
    \begin{subfigure}[b]{0.5\linewidth}
        \centering
        \includegraphics[width=\linewidth]{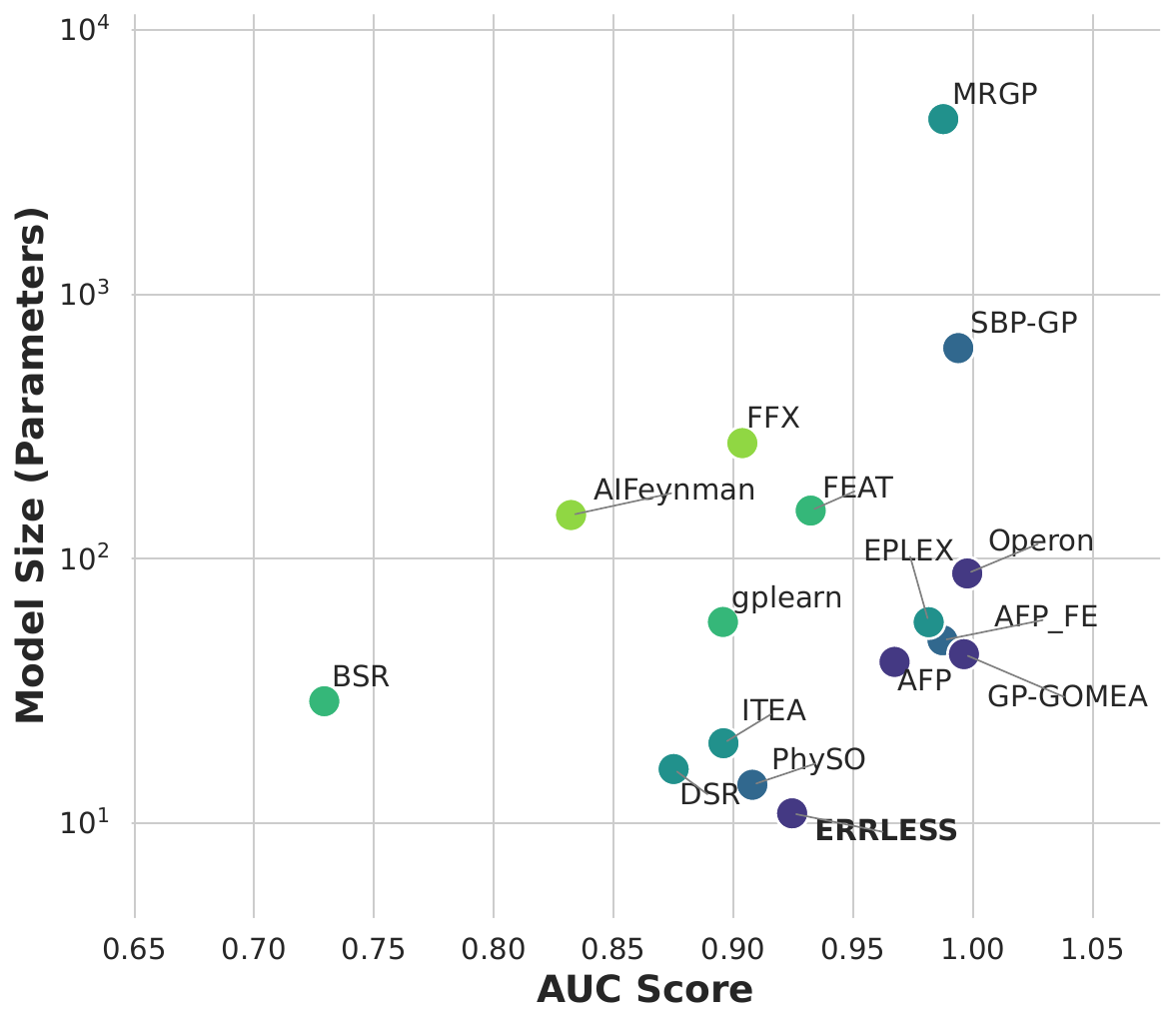}
        \caption{Accuracy vs. Complexity trade-off for $\gamma=0.01$}
        \label{fig:complexity_feynman}
    \end{subfigure}
    \vspace*{-0.5em}
    \caption{\textbf{Performance analysis on the Feynman benchmark.} (a) Robustness profile showing the AUC of the empirical success rate $\mathbb{P}[R^2\geq t]$ across varying noise levels ($\gamma \in \{0.001, 0.01, 0.1\}$). Algorithms are ranked by mean AUC, with \ourMethod\ demonstrating superior stability. (b) Pareto front illustrating the trade-off between median AUC score and model parameter count. \ourMethod\ achieves a competitive performance-to-complexity ratio compared to baselines.}
    \label{fig:ai_feynman_results}
    \vspace*{-1em}
\end{figure*}
\section{Experimental setup}
\label{setup}
In this section, we describe our experimental setup. We compare \ourMethod against the baselines from \citet{lacava2021contemprorary} and PhySO \citep{PhySO_RL_DA}, a state-of-the-art symbolic regression method that incorporates dimensional analysis constraints. A detailed description of these baselines is provided in \cref{app:baselines}. In \cref{datasets}, we briefly summarize the benchmark datasets, and in \cref{metrics}, we outline the evaluation metrics. All algorithms are allowed a budget of 1 million reward evaluations as outlined in \citet{lacava2021contemprorary}.
\subsection{Datasets}
\label{datasets}

\paragraph{Synthetic dataset.} We construct a small dataset of seven expressions to closely study the quality of posterior modeling. Each expression has $20$ points for training and $100$ points for testing. We set the maximum number of nodes to $L=9$ and the maximum number of constants to $K=3$. The expressions contain at most two variables. We give an overview of the expressions and their corresponding datasets in \cref{datasets_details}.

\paragraph{Feynman Symbolic Regression Database.} We use the Feynman Symbolic Regression Database \citep{Udrescu2020}, a standard benchmark for symbolic regression comprising 100 expressions from the \textit{Feynman Lectures on Physics} \citep{Feynman_Leighton_Sands_2015} and 20 additional physics-inspired bonus expressions. Following \citet{PhySO_RL_DA}, we remove 4 expressions involving $\arccos$ or $\arcsin$, leaving 116 expressions in total.\footnote{The expressions removed are: \texttt{feynman\_I\_26\_2}, \texttt{feynman\_I\_30\_5}, \texttt{feynman\_II\_11\_17} and  \texttt{feynman\_test\_10}.} Each expression is paired with 1 million sampled points; we subsample 10,000 for training (to compute the reward) and 25,000 for testing, as per the \texttt{SRBench} protocol \citep{lacava2021contemprorary}. We set $L=32$ and $K=3$.

The benchmark datasets are originally noise-free. Following the protocol of \citet{lacava2021contemprorary}, we add Gaussian noise to the ground-truth targets in the train set only. The noise is sampled from $\mathscr{N}\Big(0, \gamma\sqrt{\frac{1}{N}\sum_{i=1}^N y_i^2}\Big)$, where $\gamma$ controls the noise level. For each expression, we run our algorithm with five random seeds and three noise settings: $\gamma \in \{0.001, 0.01, 0.1\}$ on the same training split. Importantly, the noise is sampled once per dataset and kept fixed across all runs to ensure comparability.

\paragraph{Blackbox benchmark.} This is a set of 122 datasets intended for benchmarking machine learning methods from PMLB \citep{romano2021pmlb}. These datasets stress-test algorithms on real data where the ground-truth formula is typically unknown. We use the more challenging subset of $12$ datasets curated in \citet{aldeia2025call} and show its metadata in \cref{tab:blackbox_metadata}.

\subsection{Metrics}
\label{metrics}

Let $Y = \{y_i\}_{i=1}^N$ (resp. $\hat{Y} = \{\hat{y}_i\}_{i=1}^N$) denote the target (resp. prediction) and their mean $\bar{Y}$ (resp. $\bar{\hat{Y}}$).
\paragraph{Prediction accuracy.} We use the coefficient of determination $R^2$  between the target and prediction: \begin{equation*}
    R^2 \coloneq 1 - \frac{\sum_{i=1}^N \left(y_i - \hat{y}_i\right)^2}{\sum_{i=1}^N \left(y_i - \bar{Y}\right)^2}.
\end{equation*}
The coefficient $R^2 \leq 1$ quantifies the fraction of variance in the target $Y$ explained by the model: $R^2 = 1$ indicates a perfect fit, while predicting the mean of $Y$ gives $R^2 = 0$.%

\paragraph{AUC Score \citep{aldeia2025call}.} To quantify the robustness and predictive accuracy of each algorithm, we use a scalar metric derived from the area under the curve of its success profile. For a given algorithm evaluated over $N$ datasets across $S$ independent runs, we first determine the median prediction accuracy (expressed as the $R^2$ coefficient) for each dataset to represent its central performance. We then define the success profile as the empirical survival function, $\mathbb{P}[R^2\geq t]$, which represents the proportion of datasets where the median accuracy meets or exceeds a sliding threshold $t\in [0,1]$. The final reported metric is obtained by integrating this success profile over the unit interval. %

\section{Results}
\label{results}
\subsection{Posterior over small expressions}

To evaluate the posterior modeling accuracy, we compare \ourMethod\ against the SMC-based baseline PySIPS \citep{bomarito2025bayesian} on the synthetic datasets introduced in \cref{datasets} (noise level $0.1$). We draw $1000$ samples from the posterior for both methods. Each method is run with ten random seeds per expression. \Cref{tab:synthetic_results} shows that our approach models the posterior over expression trees more accurately, as indicated by the posterior predictive mean ($R^2_{\rm PP}$). PySIPS attains lower NLL and, on most problems, higher single-expression Test $R^2$: its search is effective at finding individual strong expressions. But its posterior-predictive mean collapses on most problems, because a tail of posterior samples diverges outside the training region and dominates the mean (cf.\ \cref{fig:posterior_plot}). This produces the large negative $R^2_{\rm PP}$ values for PySIPS in \cref{tab:synthetic_results}. \ourMethod\ is less prone to this failure: its posterior draws yield a finite predictive mean on every problem. The example in \cref{fig:posterior_plot} shows that \ourMethod\ produces posterior samples that fit the data even though the training region is too small to reveal the true shape of the function. %

\subsection{Fit quality}

\ourMethod\ achieves a competitive AUC score of 0.924 on the Feynman database, maintaining superior robustness across all noise levels compared to closely related reinforcement learning baselines such as DSR (0.873) and PhySO (0.893), the latter of which incorporates dimensional analysis constraints. As illustrated in \cref{fig:ai_feynman_results}, our approach is stable, with almost no variation in AUC score as the noise level increases. \ourMethod\ also outperforms the Bayesian symbolic regression approach, BSR (0.702). The performance gap between \ourMethod\ and leading methods, such as GP-GOMEA and Operon, can be attributed to two key factors: (1) unlike most competing methods that optimize directly for reward, \ourMethod\ must approximate the joint posterior over expressions, constants, and noise, a comprehensive probabilistic task that typically necessitates a larger training budget than the 1M evaluations prescribed by the SRBench framework; and (2) while leading methods often achieve high fit scores, they frequently produce overly complex expressions containing hundreds of constants \citep{PhySO_RL_DA} (see \cref{fig:complexity_feynman}). Such results often defeat the primary objective of symbolic regression, which is to identify parsimonious and interpretable expressions.

\paragraph{Blackbox benchmark.} From \cref{fig:blackbox_results}, \ourMethod reaches an AUC of 0.350, which is higher than BSR (0.19) and AIFeynman (0.004) but lower than XGBoost (0.625). This benchmark is difficult because some datasets have a large number of data points, making the likelihood very peaky and hard to navigate for the policy. However, \ourMethod remains competitive in terms of striking a balance between the accuracy and complexity of the generated expressions.

An additional benefit of \ourMethod is its speed: we show in \cref{fig:runtime} that it is an order of magnitude faster than most learning-based competitors, which can be attributed to the fact that \ourMethod does not require an inner loop to optimize the constants, the typical bottleneck for SR algorithms.

\subsection{Qualitative analysis}
\looseness=-1
We examine expressions discovered by our sampler and highlight the case of the ground-truth expression $\sfrac{\rho_0}{\sqrt{1 - \tfrac{v^2}{c^2}}}$. The highest-scoring candidate found by \ourMethod\ was $\tfrac{\rho_0}{\cos(v/c)}$. Although these two expressions appear unrelated at first glance, their denominators have the same second-order Taylor approximation in $\frac vc$ at 0. The length of the postorder representation of the ground-truth expression is greater than that of its approximate counterpart, which is unfavored by the prior. An extreme example of the same is the ground-truth expression $\sfrac{0.159 h\omega}{\left(\exp\Big(0.159\tfrac{h\omega}{Tk_B}\Big) - 1\right)}$, which has the leading-order Taylor approximation $Tk_B$. The prior strongly favors $Tk_B$, and the two expressions have similar likelihood: the linear approximation achieves a test set $R^2$ of $0.99$. 

These results can be explained by the fact that \ourMethod\ incorporates a prior that favors concise expressions, as opposed to methods that only optimize the quality of fit. With more data samples, we would expect the likelihood to eventually dominate the prior in the reward, and the ground-truth expression would have a higher score.

\section{Conclusion}
\label{conclusion}
\raggedbottom
We have introduced \ourMethod, a scalable approach to Bayesian symbolic regression, using maximum-entropy reinforcement learning to amortize posterior sampling over algebraic expressions describing a stochastic dependence of a target variable on its inputs. We formulate expression synthesis as a sequential decision-making process and prune the search space by enforcing dimensional constraints during bottom-up construction. On the Feynman Symbolic Regression Database~\citep{Udrescu2020}, \ourMethod achieves a competitive AUC score while approximating the full posterior distribution, rather than returning a single-point estimate.

\paragraph{Limitations and future work.} While \ourMethod accurately models the posterior over expressions, performance can degrade on highly complex target expressions. Future work could condition the sampler on the temperatures of the log-likelihood and the priors to better trade off complexity and accuracy. Additionally, amortizing the sampler over datasets would allow posterior sampling for new datasets without the need for retraining. Symbolic regression methods that amortize over datasets have already shown promise in optimizing for the best expression given a dataset~\citep{biggio2021neural, kamienny2023deep}. This naturally motivates extending \ourMethod to learn the operator library itself, allowing for reusable constructs that recur across datasets~\citep{ellis2021dreamcoder}. Finally, extending our work to discovering symbolic forms of partial and ordinary differential equations would be a strong future direction given their prevalence in science.

\begin{acknowledgements} %
O. Boussif was supported by the National Research Council Canada (NRC)
AI4D program. 
M. Jain is supported by an FRQNT Doctoral Fellowship (\url{https://doi.org/10.69777/366694}). 
Y.~Kaddar was supported by the Oxford--DeepMind Graduate Scholarship.
E.S.\ Whitammer acknowledges support from the CIFAR Learning in Machines and Brains Programme.
The research was also enabled by computational resources provided by the Digital Research Alliance of Canada (\url{https://alliancecan.ca}) and Mila (\url{https://mila.quebec}).
\end{acknowledgements}

\bibliography{uai2026-template}

\newpage

\onecolumn

\title{Bayesian Symbolic Regression with Entropic Reinforcement Learning\\(Supplementary Material)}
\maketitle

\appendix
\section{Bottom-up generation}
\subsection{Operator physical unit constraints and assignments}
\label{unit_constraints}
We show the values of the partial \emph{unit assignment} function for the operators that we use in our library in \cref{tab:unit_functions_full}.

\begin{table}[h!]
\centering
\caption{Physical unit assignment function for the full operator library. When a variable is dimensionless, its unit vector is $\mathbf{0}$.}
\label{tab:unit_functions_full}
\begin{tabular}{cc}
\toprule
\textbf{Operator $g$} & \textbf{Unit assignment function $\mathcal{U}_g$}  \\
\midrule
$+$ & $\mathcal{U}_+(\vu_1, \vu_2) = \vu_1$ if $\vu_1=\vu_2$, else undefined  \\
$-$ & $\mathcal{U}_-(\vu_1, \vu_2) = \vu_1$ if $\vu_1=\vu_2$, else undefined  \\
$\times$ & $\mathcal{U}_\times(\vu_1, \vu_2) = \vu_1 + \vu_2$ \\
$/$ & $\mathcal{U}_/(\vu_1, \vu_2) = \vu_1 - \vu_2$ \\
\midrule
$\sin$ & $\mathcal{U}_{\sin}(\vu) = \mathbf{0}$  if $\vu=\mathbf{0}$, else undefined \\
$\cos$ & $\mathcal{U}_{\cos}(\vu) = \mathbf{0}$  if $\vu=\mathbf{0}$, else undefined \\
$\log$ & $\mathcal{U}_{\log}(\vu) = \mathbf{0}$  if $\vu=\mathbf{0}$, else undefined \\
$\exp$ & $\mathcal{U}_{\exp}(\vu) = \mathbf{0}$  if $\vu=\mathbf{0}$, else undefined \\
$\square^2$ & $\mathcal{U}_{\square^2}(\vu) = 2 \vu$ \\
$\sqrt{\square}$ & $\mathcal{U}_{\sqrt{\square}}(\vu) =\frac{1}{2} \vu$ \\
$-\square$ & $\mathcal{U}_{-\square}(\vu) =  \vu$ \\\bottomrule
\end{tabular}
\end{table}

\subsection{Reward function}
\label{unigram}
\paragraph{Unigram prior over expressions.} Following \citet{constantin2024statistical}, the frequency of mathematical operators in physics equations decays exponentially with rank. We leverage this observation by constructing a unigram prior over operators based on their empirical frequencies. Specifically, we use \emph{Encyclopaedia Inflationaris} as the reference corpus, extract operator frequencies (Table 1 in their paper), discard operators not included in our library, and renormalize the distribution. The resulting frequencies for our operator set are shown in \cref{unigram_prior}.
\begin{table}[h]
\centering
\caption{Unigram prior over operators, variables, and constants. Frequencies are normalized after restricting to our operator library.}
\label{unigram_prior}
\begin{tabular}{@{}l c}
\toprule
\textbf{Token} & \textbf{Frequency} \\
\midrule
$\times$  & 0.1770 \\
$/$       & 0.1328 \\
$-$       & 0.0476 \\
$+$       & 0.0454 \\
$\square^2$ & 0.0365 \\
$\exp$    & 0.0210 \\
$\sqrt{\square}$ & 0.0199 \\
$-\square$ & 0.0177 \\
$\log$    & 0.0133 \\
$\cos$    & 0.0072 \\
$\sin$    & 0.0048 \\
\midrule
Variables & 0.2877 \\
Constants & 0.1892 \\
\bottomrule
\end{tabular}
\end{table}

\paragraph{Prior over constant assignments} For the \textbf{synthetic} dataset we use a Gaussian with mean $0$ and standard deviation $10$ whereas for the \textbf{Feynman} and \textbf{Blackbox} datasets the standard deviation is $20$.

\paragraph{Prior over the noise $\sigma$} For the \textbf{synthetic} dataset we use $\log\sigma\sim\mathcal{N}(0,5^2)$, \ie\ $\sigma\sim\operatorname{LogNormal}(0,5)$, except for $\sqrt{x^2+y^2}$, where $\log\sigma\sim\mathcal{N}(-1,5^2)$; for the \textbf{Feynman} and \textbf{Blackbox} datasets we use a Half-Normal with scale parameter $2000$.

\paragraph{Posterior-predictive NLL estimator}\label{app:nll-estimator}
The NLL column of \cref{tab:synthetic_results} is computed identically for both methods as
\[
\mathrm{NLL} \;=\; -\sum_{i=1}^{N}\log\frac{1}{J}\sum_{j=1}^{J}\mathcal{N}\!\big(y_i;\, f_{\T_j,\vtheta_j}(\vx_i),\, \sigma_j^2\big),
\]
where $\{(\vx_i, y_i)\}_{i=1}^{N}$ is the test set, $(\T_j,\vtheta_j)$ is the $j$-th posterior draw, $J$ is the number of retained draws (up to $1000$; draws with non-finite predictions are discarded), and $\sigma_j$ is the $j$-th draw's sampled noise value for \ourMethod\ and the $j$-th particle's root-mean-square residual on the training split for PySIPS.

\section{Experimental details}
\label{exp_details}
\subsection{Datasets}
\label{datasets_details}
\paragraph{Synthetic.} \cref{tab:synthetic} shows the seven expressions we used for the benchmark alongside the ranges used for the training and test set, respectively.

\begin{table}[H]
\centering
\caption{\textbf{Synthetic dataset}. Variables are sampled uniformly from the specified intervals.}
\begin{tabular}{@{}lcc}
\toprule
\textbf{Expression} & \textbf{Training range} & \textbf{Test range} \\
\midrule
$1/\sqrt{1-x^2}$ & $x \sim \mathscr{U}(-0.6,\,0.6)$ & $x \sim \mathscr{U}(-0.99,\,0.99)$ \\
$2.37x+3.02$ & $x \sim \mathscr{U}(-0.5,\,0.5)$ & $x \sim \mathscr{U}(-1,\,1)$ \\
$x + \sin(5.5x)$ & $x \sim \mathscr{U}(-0.5,\,0.5)$ & $x \sim \mathscr{U}(-2,\,2)$ \\
$4.567\exp{x}+y$ & $x,y \sim \mathscr{U}(-1,\,1)$ & $x,y \sim \mathscr{U}(-2,\,2)$ \\
$\sin(y + 2.5x)$ & $x,y \sim \mathscr{U}(0,\,1)$ & $x,y \sim \mathscr{U}(0,\,3)$ \\
$\sqrt{x^2 + y^2}$ & $x,y \sim \mathscr{U}(0,\,1)$ & $x,y \sim \mathscr{U}(0,\,3)$ \\
$\sin(x)\cos(y)$ & $x,y \sim \mathscr{U}(0,\,1)$ & $x,y \sim \mathscr{U}(0,\,3)$ \\
\bottomrule
\end{tabular}
\label{tab:synthetic}
\end{table}

\begin{figure}[H]
        \centering
        \includegraphics[width=0.99\linewidth]{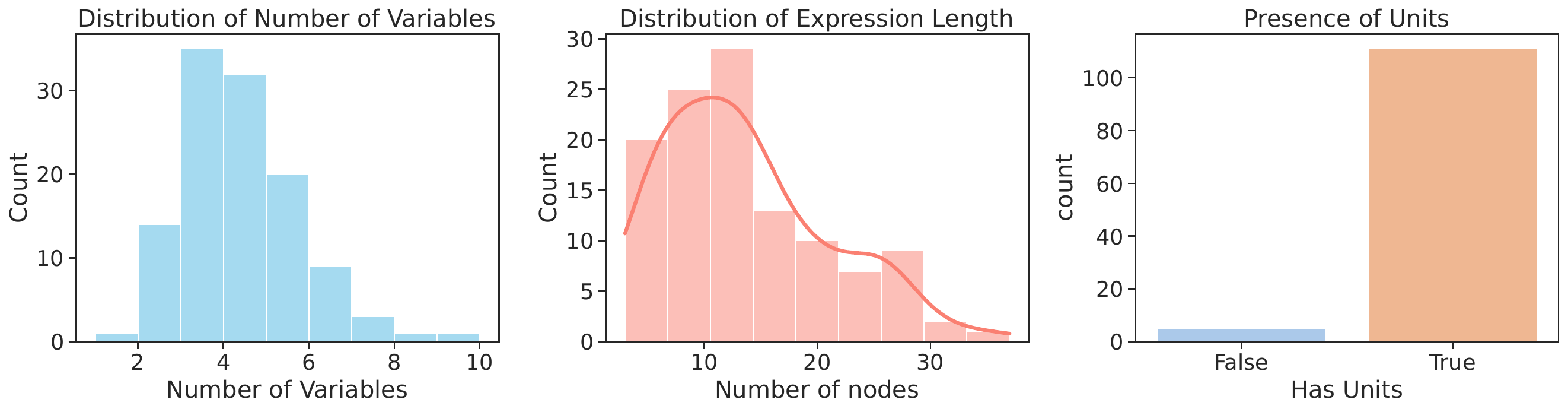}  
        \caption{\textbf{Exploratory analysis of the Feynman Symbolic Regression Database}. Left: histogram of the number of variables per expression. Center: distribution of expression tree lengths. Right: proportion of expressions in which all variables are associated with physical units.}
        \label{fig:ai_feynman}
\end{figure}

\paragraph{Feynman Symbolic Regression Database.} We show characteristics of the Feynman Symbolic Regression Database \citep{Udrescu2020} (see \cref{fig:ai_feynman}). In particular: (i) the distribution over the number of variables in a given expression in the dataset, (ii) the distribution over expression lengths of a given expression in the dataset, and (iii) the ratio of expressions with physical units compared to unitless ones. 

\paragraph{Blackbox benchmark.} We show the metadata of the benchmark in \cref{tab:blackbox_metadata} where for each dataset we show the number of its datapoints as well as the number of features. 

\begin{table}
\caption{\textbf{Blackbox benchmark metadata.} We show the dataset size and number of features of each dataset in the Blackbox benchmark.}

\centering
\begin{tabular}{@{}lrr}
\toprule
\textbf{Dataset} & \textbf{Size} & \textbf{Number of features} \\
\midrule
\texttt{1028\_SWD} & $1000$ & $11$ \\
\texttt{1089\_USCrime} & $47$ & $14$ \\
\texttt{1193\_BNG\_lowbwt} & $31104$ & $10$ \\
\texttt{1199\_BNG\_echoMonths} & $17496$ & $10$ \\
\texttt{192\_vineyard} & $52$ & $3$ \\
\texttt{210\_cloud} & $108$ & $6$\\
\texttt{522\_pm10} & $500$ & $8$ \\
\texttt{557\_analcatdata\_apnea1} & $475$ & $4$ \\
\texttt{579\_fri\_c0\_250\_5} & $250$ & $6$ \\
\texttt{606\_fri\_c2\_1000\_10} & $1000$ & $11$ \\
\texttt{650\_fri\_c0\_500\_50} & $500$ & $51$ \\
\texttt{678\_visualizing\_environmental} & $111$ & $4$ \\
\bottomrule
\end{tabular}
\label{tab:blackbox_metadata}
\end{table}

\subsection{Replay buffer}
We use a modified version of the prioritized replay buffer where each expression tree $\T$ can be stored up to a maximum of $N_{\text{repeat}}$ times in the buffer. This is to ensure that the model is trained on different values of the constants for the same expression. When a new batch of elements is added, we apply a first filtering step where we only keep the ones that have a reward higher than the minimal reward in the buffer. This ensures that the minimum reward in the buffer never decreases, and that we do not hinder the quality of the samples in the buffer. After that, we perform a second filtering step where we make sure that none of the added expression trees are repeated more than $N_{\text{repeat}}$ times.

\subsection{Hyperparameters}
\paragraph{Policy.} Recall that an expression tree $\T$ is uniquely represented by its postorder sequence $\mathcal{W}(\T)\in\Sigma^*$. We first embed it using a learnable embedding matrix with hidden dimension $256$ to get a representation. A learned positional embedding is then added to this representation, which is processed by a Transformer encoder with $2$ layers and $4$ attention heads \citep{attention}. The output of the encoder is used in three ways:

\begin{itemize}[left=0pt,nosep]
    \item Action selection: A linear layer maps the representation to logits over the discrete action space (operators, variables, and termination).
    \item Constant generation: We approximate the generative distribution of the constants using a Gaussian Mixture Model (GMM) with $5$ components. The latent representation produced by the transformer is processed by three distinct linear heads: (1) one to parameterize the means of each Gaussian component, (2) one to determine the corresponding covariances, and (3) one to predict the mixture weights.
    \item Noise generation: In a manner analogous to constant generation, the noise distribution is approximated by a mixture model. For this task, we use a Log-Normal mixture to ensure the generated noise parameters remain within a valid, positive-valued domain. The number of components in the mixture is also set to $5$.
\end{itemize}
\paragraph{Off-policy training.} We use an $\epsilon$-greedy policy where $\epsilon$ decays linearly from $1$ to $0.05$ midway through the training and then for the other half it stays at $0.05$. We use a prioritized replay buffer with a capacity of $10000$ trajectories and $N_{\text{repeat}}=3$. The batch size is set to $800$ and the ratio of samples coming from the replay buffer decays linearly from $0.9$ to $0.2$ (see \cref{sec:ablation_mixing} for ablation). We train the policy for a total of $1250$ iterations, which corresponds to $1$M reward function calls.
\paragraph{Reward function.} The inverse temperature of the prior factors is $\alpha=1$ for the expression tree log-prior and $\lambda=1$ for the parameter log-prior.
\paragraph{Optimization.} We use the Adam optimizer without weight decay \citep{kingma2014adam0}. The policy $\log \pi_\varphi(\T,\vtheta,\sigma)$ learning rate is set to $0.0001$ and that of the log-partition function $\log Z_\varphi$ to $0.01$.

\subsection{Baselines}\label{app:baselines}

\paragraph{AFP / AFP\_FE \citep{schmidtAgefitnessParetoOptimization2011,schmidt2009distilling}.}

Age-Fitness Pareto optimization (AFP) is a genetic programming (GP) strategy that frames search as a bi-objective problem over prediction error and individual age. Each candidate solution is assigned a fitness value and an age, defined as the number of generations since creation. At each generation, selection operates on the Pareto front with respect to these two objectives, favoring individuals that are either accurate and relatively young or novel relative to the current population. By rewarding both accuracy and youth, AFP maintains diversity and reduces premature convergence, while still steering the search toward low-error expressions.

\paragraph{AIFeynman \citep{Udrescu2020}.}

AIFeynman is a physics-inspired, multi-stage method designed to discover symbolic expressions by systematically breaking down a complex problem into simpler ones. It does not learn a single generative model but rather follows a deterministic, divide-and-conquer strategy. First, a neural network is trained to high accuracy on the dataset $\D$. This network is then treated as an ``oracle'' and is probed to discover properties of the underlying function, such as symmetries, separability, or polynomial structure. Based on these discovered properties, the original problem is recursively simplified. The final, simplified sub-problems are then solved using a combination of brute-force search and polynomial fitting.

\paragraph{BSR \citep{jin2019bayesian}.}

Bayesian symbolic regression (BSR) directly addresses the problem of posterior inference over the space of expressions. Similar to our approach, BSR also aims to sample from the posterior distribution $p(\T, \vtheta \mid \D)$, where $\T$ is the expression tree structure and $\vtheta$ represents its constant parameters. Due to the varying dimensionality of its parameter space (due to constantly changing tree structure), BSR employs Reversible-Jump MCMC (RJMCMC), a technique that allows the MCMC sampler to propose ``moves'' between models of different dimensions (\eg, adding or removing a node in the tree $\T$) while maintaining the detailed balance condition. This approach relies on handcrafted proposal distributions for these moves, so it can be computationally intensive and often fails to explore the posterior landscape sufficiently.

\paragraph{DSR \citep{petersen2019deep}.}

Deep symbolic regression (DSR) was a seminal work that first framed the symbolic regression task as a reinforcement learning (RL) problem. DSR employs a Recurrent Neural Network (RNN) as a policy, which autoregressively generates an expression tree $\T$ token by token in a top-down, pre-order traversal. A complete expression constitutes a trajectory, and the quality of this expression (\eg, its $R^2$ score on the dataset $\D$) serves as the reward $R(\T)$. The policy is trained using a risk-seeking policy gradient algorithm, which biases the search towards high-reward expressions and helps escape local optima. The ultimate goal of DSR is to find a single best-fitting expression that maximizes the expected reward.
\paragraph{PhySO \citep{PhySO_RL_DA}.} PhySO uses the same learning framework as DSR for training its policy, but it adds physical unit constraints where it forces the expression to be dimensionally valid at each generation step.

\paragraph{EPLEX \citep{lacavaProbabilisticMultiobjectiveAnalysis2019}.}

EPLEX is an advanced parent selection method for Genetic Programming designed to excel in continuous-valued regression tasks. Traditional selection methods rely on an aggregate fitness score (like average error), which loses information about performance on individual data points. EPLEX addresses this by filtering the population sequentially on a random ordering of individual training cases.

\paragraph{FEAT \citep{lacavaLearningConciseRepresentations2019}.}

FEAT is a hybrid method for symbolic regression that combines evolutionary computation with linear models. Instead of evolving a single monolithic expression, FEAT evolves a set of simpler expression trees that serve as features. These features are then used as inputs to a linear model. The key innovation is a feedback mechanism where the coefficients learned by the linear model are used to guide the evolutionary search, prioritizing the mutation and replacement of less impactful features.

\paragraph{FFX \citep{mcconaghyFFXFastScalable2011}.}

FFX is a non-evolutionary, deterministic algorithm for symbolic regression that casts the problem as a feature selection task within a generalized linear model. The method operates in two main stages: first, it deterministically generates a massive library of candidate basis functions by applying a predefined set of nonlinear operators and interactions to the input variables. Second, it employs path-wise regularized learning (specifically, an elastic net) to efficiently search this vast feature space.

\paragraph{GP-GOMEA \citep{virgolin2020improving}.}

GP-GOMEA is a model-based evolutionary algorithm that aims to improve search efficiency by explicitly learning and exploiting the structure of promising solutions. Unlike traditional GP, which relies on blind genetic operators like crossover and mutation, GP-GOMEA learns a ``linkage model'' in each generation. This model, typically a Linkage Tree built using mutual information between nodes in the population's expression trees, identifies groups of genes (sub-programs) that work well together.

\paragraph{gplearn \citep{gplearn}.}

\texttt{gplearn} is a Python library that implements tree-based genetic programming for symbolic regression within the \texttt{scikit-learn} API \citep{sklearn_api}. Candidate models are mathematical expression trees evolved using standard GP operators such as sub-tree crossover and mutation, with fitness measured by prediction error.

\paragraph{ITEA \citep{defrancaInteractionTransformationEvolutionaryAlgorithm2020}.} 

ITEA is an evolutionary algorithm that operates on a constrained representation called Interaction Transformation (IT). Unlike the free-form trees in traditional GP, an IT expression is restricted to a linear combination of nonlinear terms. ITEA evolves a population of these structured expressions using a mutation-only strategy.

\paragraph{MRGP \citep{arnaldoMultipleRegressionGenetic2014a}.}

MRGP is a hybrid technique that integrates tree-based GP with the LASSO regularization method. Unlike conventional GP, MRGP does not directly compare the final program's output with the target variable. Instead, it constructs a set of sub-expressions from the program and fits a linear combination of these sub-expressions to the target output. The target variable is then compared against the output of the resulting regression model.

\paragraph{Operon \citep{kommendaParameterIdentificationSymbolic2019}.}

Operon is a modern, highly efficient C++ framework for GP in symbolic regression. It focuses on improving performance and scalability through advanced architectural choices, including representing expression trees in a cache-friendly, continuous memory layout. It also implements a fine-grained, low-overhead concurrency model for parallel execution.

\paragraph{SBP-GP \citep{virgolinLinearScalingSemantic2019}.}

SBP-GP is a GP method that guides variation using semantic backpropagation (SB). Instead of random tree changes, SB computes the output each sub-tree should produce to help the overall expression match the target, by propagating the target values down through the tree using function inverses. New sub-trees are then generated or selected to better match these desired outputs.

\section{Posterior predictive}
\label{posterior_predictive}

For any test input $\vx^\ast$, the posterior predictive distribution of the corresponding output $\ry^\ast$ is given by:
\begin{equation}
    p(\ry^*\mid\vx^*, \D) = \sum_{\T}\int p(\ry^*\mid\vx^*,\T,\vtheta)\,p(\T,\vtheta\mid \D)\,\dd\vtheta 
\end{equation}
Intuitively, this distribution weighs the contribution of all expression trees and their parameters. 
\paragraph{Mean of the posterior predictive.} In general, the mean is:
\begin{equation*}
\mathbb{E}\left[\ry^*\mid\vx^*, \D\right] = \sum_T \int p(\T,\vtheta\mid \D)\mathbb{E}\left[\ry^*\mid\vx^*,\T,\vtheta\right] \dd\vtheta
\end{equation*}

Since the likelihood model is a Gaussian (\ref{conditional-rv}), we have $\mathbb{E}\left[\ry^*\mid\vx^*,\T,\vtheta\right]=f_{\T,\vtheta}(\vx^*)$ which yields:
\begin{equation}
\mathbb{E}\left[\ry^*\mid\vx^*, \D\right] = \mathbb{E}_{\T,\vtheta}\left[f_{\T,\vtheta}(\vx^*)\right]
\end{equation}
In practice, we compute the above quantity by drawing expression trees with their parameters from the learned approximation to the posterior. A tempered posterior can be obtained by weighting these samples by their probability under the trained sampler raised to a power. For example, for a posterior at temperature $\frac12$, we simply reweight the samples by their likelihood under the trained model.

\begin{figure*}[t]
    \centering
    \vspace*{-5mm}
    \begin{subfigure}[b]{0.35\linewidth}
        \centering
        \includegraphics[width=\linewidth]{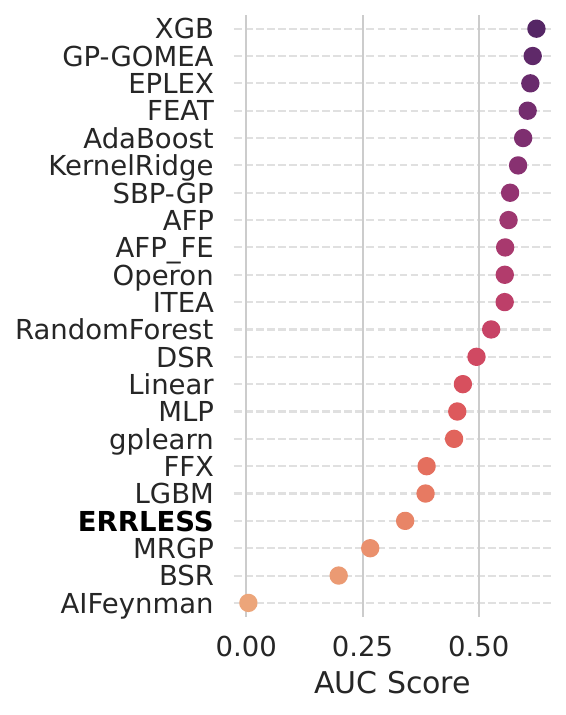}
        \caption{AUC Score on the Blackbox benchmark}
        \label{fig:blackbox}
    \end{subfigure}
    \hfill
    \begin{subfigure}[b]{0.5\linewidth}
        \centering
        \includegraphics[width=\linewidth]{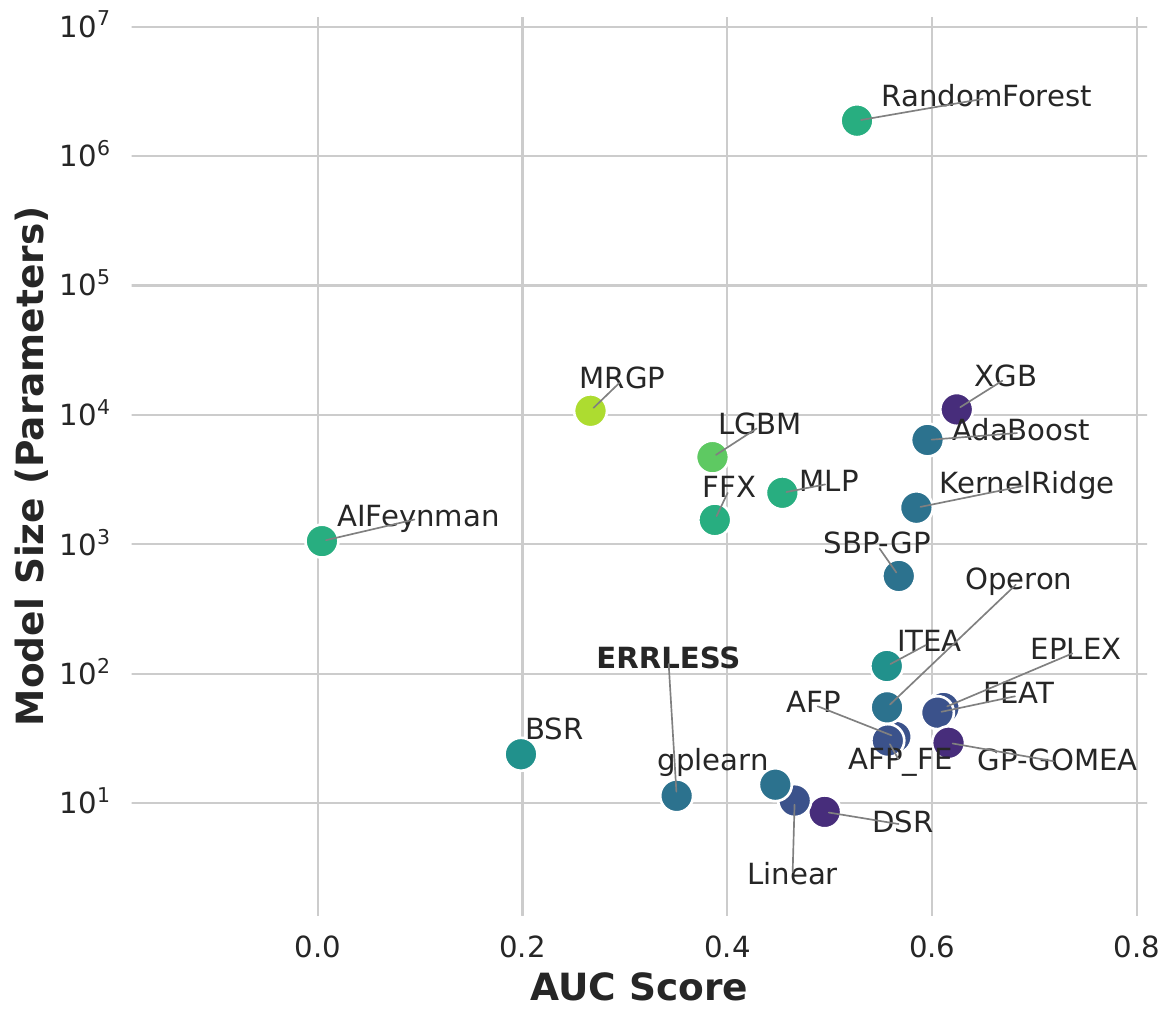}
        \caption{Accuracy vs. Complexity trade-off}
        \label{fig:complexity_blackbox}
    \end{subfigure}
    \caption{\textbf{Performance analysis on the Blackbox benchmark.} (a) Area Under the Curve (AUC) of the empirical success rate $\mathbb{P}[R^2\geq t]$. Algorithms are ranked by mean AUC. (b) Pareto front illustrating the trade-off between median AUC score and model parameter count. \ourMethod\ achieves a competitive performance-to-complexity ratio compared to baselines.}
    \label{fig:blackbox_results}
    \vspace{-6mm}
\end{figure*}


\section{Ablation results}
We benchmark the design choices by ablating the expression prior and the objective and by running a sensitivity analysis on the mixing coefficient of the replay buffer on the synthetic dataset (see \cref{datasets_details}). In all ablation tables, each configuration is run with five random seeds per expression: model size is computed for each expression as the median over seeds and then averaged over the synthetic expressions.

\subsection{Expression prior}
\label{sec:ablation_prior}
We run \ourMethod with different choices of the expression prior $P(\T)$. Because the learning problem is invariant to multiplication of the reward by a scalar, one can define priors by choices of the term $P(\T)$ that do not necessarily sum to 1 over $\T\in\mathcal{T}_\Sigma$. Such unnormalized choices are well-defined: the maximum number of nodes is bounded in all our experiments, so the set of reachable expression trees is finite and every positive $P(\T)$ restricted to it induces a proper prior after normalization.
\begin{itemize}
    \item \textbf{Unigram prior}: This is the original expression prior used for the main results in the paper; see \cref{unigram}.
    \item \textbf{Uniform prior}:  $\log P(\T)=0$.
    \item \textbf{PCFG prior}: A probabilistic context-free grammar prior fit on the Wikipedia named equations and Encyclopaedia Inflationaris \citep{constantin2024statistical}.
    \item \textbf{Nodes prior}: This is a prior that penalizes the number of nodes in the expression and is given by $\log P(\T) = -n(\T)$, where $n(\T)$ is the number of nodes in $\T$.
\end{itemize}

\begin{table}[H]
\centering
\caption{Ablation results on the expression prior for different noise levels.}
\begin{tabular}{@{}llcc}
\toprule
\textbf{Noise level} & \textbf{Prior} & \textbf{AUC score} & \textbf{Model size} \\
\midrule
\multirow{4}{*}{0.001} & Nodes Prior    & 0.817349 & 7.857143 \\
                      & PCFG Prior     & 0.799313 & 5.857143 \\
                      & Uniform Prior     & 0.822502 & 8.142857 \\
                      & Unigram Prior  & 0.818208 & 6.142857 \\
\cmidrule(l){2-4}
\multirow{4}{*}{0.010} & Nodes Prior    & 0.817349 & 7.857143 \\
                      & PCFG Prior     & 0.770684 & 5.428571 \\
                      & Uniform Prior       & 0.817349 & 7.857143 \\
                      & Unigram Prior  & 0.776410 & 5.428571 \\
\cmidrule(l){2-4}
\multirow{4}{*}{0.100} & Nodes Prior    & 0.895076 & 6.142857 \\
                      & PCFG Prior     & 0.548382 & 5.428571 \\
                      & Uniform Prior       & 0.579588 & 10.000000 \\
                      & Unigram Prior  & 0.778414 & 5.285714 \\
\bottomrule
\end{tabular}
\label{tab:noise_prior_results}
\end{table}

The Unigram prior performs consistently well across all noise levels (\cref{tab:noise_prior_results}). The Nodes prior excels in high-noise settings, where the Uniform prior struggles despite generating longer expressions. Similarly, the PCFG prior degrades as noise increases.

\subsection{Training objective}
\label{sec:ablation_objective}
In this section we ablate the maximum-entropy RL training objective and compare trajectory balance (TB), which we used for the main results, against detailed balance (DB), equivalent in our setting to optimization of a soft Bellman error as in \citet{haarnoja2018soft}, cf.\ \citet{tiapkin2024generative,deleu2024discrete}.

Recall from \cref{bottomup} that an expression tree $\T$ is uniquely represented by its postorder representation $w_1\dots w_n \top$. DB requires learning a scalar value (log-flow) function $\log F_\varphi (w_1\dots w_t)$ (where $t\leq n+1$ and $w_{n+1}=\top$). The objective associated with $(\T,\vtheta,\sigma)$ is:

\begin{equation}
    \mathcal{L}_{\text{DB}}(\T, \vtheta, \sigma;\varphi) \coloneq \big[\log F_\varphi (w_1\dots w_t) + \log \pi_\varphi(w_{t+1}|w_1\dots w_t) - \log F_\varphi (w_1\dots w_{t+1})\big]^2,
    \label{eq:db}
\end{equation}
where we impose that $F_\varphi(w_1\dots w_n \top) = R(\T,\vtheta,\sigma)$. 

\begin{table}[H]
\centering
\caption{Ablation results on the training objective for different noise levels.}
\begin{tabular}{@{}llcc}
\toprule
\textbf{Noise level} & \textbf{Objective} & \textbf{AUC score} & \textbf{Model size} \\
\midrule
\multirow{2}{*}{0.001} & DB & 0.844260 & 7.714286 \\
                       & TB & 0.818208 & 6.142857 \\
\cmidrule(l){2-4}
\multirow{2}{*}{0.010} & DB & 0.786859 & 6.857143 \\
                       & TB & 0.776410 & 5.428571 \\
\cmidrule(l){2-4}
\multirow{2}{*}{0.100} & DB & 0.748354 & 7.428571 \\
                       & TB & 0.778414 & 5.285714 \\
\bottomrule
\end{tabular}
\label{tab:obj_noise_results}
\end{table}

From \cref{tab:obj_noise_results}, we observe that the difference in AUC is inconclusive, but that TB tends to produce shorter expressions on average, especially as the noise level increases. We note, however, that DB requires an additional flow network, which adds parameter count and consumes more resources.

\subsection{Replay buffer mixing coefficient}
\label{sec:ablation_mixing}
We perform a sensitivity analysis where we fix the mixing coefficient (instead of annealing it from 0.9 to 0.2 as we do for the main results) on the synthetic dataset for noise level 0.01:

\begin{table}[H]
\centering
\caption{Sensitivity analysis for the replay buffer mixing coefficient at noise level 0.01.}
\begin{tabular}{@{}lcc}
\toprule
\textbf{Mixing coefficient} & \textbf{AUC score} & \textbf{Model size} \\
\midrule
0.1 & 0.853851 & 7.857143 \\
0.2 & 0.817349 & 7.428571 \\
0.3 & 0.813914 & 7.857143 \\
0.4 & 0.960063 & 6.857143 \\
0.5 & 0.947466 & 7.571429 \\
0.6 & 0.817349 & 7.428571 \\
0.7 & 0.952906 & 7.142857 \\
0.8 & 0.960063 & 6.857143 \\
0.9 & 0.817349 & 7.857143 \\
\bottomrule
\end{tabular}
\label{tab:mixing_coeff}
\end{table}

The best performance is obtained for mixing coefficients of 0.4 and 0.8 (\cref{tab:mixing_coeff}). More critically, this shows that fixing the mixing coefficient works better than annealing in this case. 

\clearpage
\section{Computational runtime}
\begin{figure}
     \centering
     \includegraphics[width=0.6\linewidth]{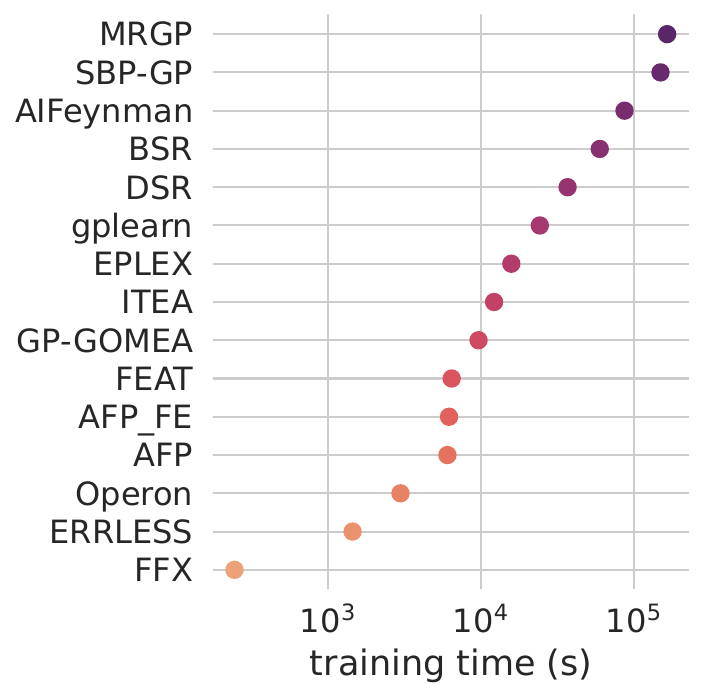}
     \caption{{\textbf{Computational runtime.} Average of the time (in seconds) taken by all methods over random seeds and expressions in the Blackbox benchmark.}}
     \label{fig:runtime}
 \end{figure}
\cref{fig:runtime} reports the runtime (in seconds) for all methods evaluated on the blackbox dataset. PhySO is omitted because its authors do not provide runtime measurements. Baseline runtimes are taken from SRBench \citep{lacava2021contemprorary}. \ourMethod was run on a machine equipped with an L40S GPU, 4 CPUs, and 48 GB of RAM.

From the figure, we observe that our method is among the fastest. While hardware differences and code optimization must be considered for a fully fair comparison, we emphasize that, unlike the baselines, our approach avoids a major bottleneck: parameter optimization for expressions. A key advantage of our method is that it learns a posterior over both expressions and parameters, eliminating the need for computationally expensive optimization procedures.

\end{document}